\pdfoutput=1

\documentclass[11pt]{article}

\usepackage[]{EMNLP2022}

\usepackage{times}
\usepackage{latexsym}

\usepackage[T1]{fontenc}

\usepackage[utf8]{inputenc}

\usepackage{inconsolata}

\usepackage{amssymb}
\usepackage{amsmath}
\usepackage{booktabs} 

\usepackage{physics}

\usepackage{graphicx}

\usepackage{amsthm}

\usepackage{color}
\usepackage{lipsum}
\usepackage{colortbl}

\def\parsum#1{\bgroup \textcolor{blue}{Paragraph summary: #1}\egroup}
\def\sectionsum#1{\bgroup \textcolor{green}{Section content: #1}\egroup \\}

\usepackage{xurl}
\usepackage{multirow}

\title{One size does not fit all: Investigating strategies for differentially-private learning across NLP tasks}

\author{Manuel Senge\thanks{~~Equal contribution}
	\and Timour Igamberdiev\footnotemark[1]
	\and Ivan Habernal	\\
	Trustworthy Human Language Technologies \\
	Department of Computer Science \\
	Technical University of Darmstadt \\
	\texttt{manuel.senge@web.de} \\ \texttt{\{timour.igamberdiev, ivan.habernal\}@tu-darmstadt.de}\\
	\url{www.trusthlt.org}
}

\begin{document}

\onecolumn
\noindent \textbf{One size does not fit all: Investigating strategies for differentially-private learning across NLP tasks}

\medskip
\noindent Manuel Senge, Timour Igamberdiev, Ivan Habernal

\bigskip
This is a \textbf{camera-ready version} of the article accepted for publication at \emph{EMNLP 2022}. The final official version will be published on the ACL Anthology website in late 2022: \url{https://aclanthology.org/}

\medskip
Please cite this pre-print version as follows.
\medskip

\begin{verbatim}
@InProceedings{Senge.et.al.2022.EMNLP,
    title = {{One size does not fit all: Investigating strategies
              for differentially-private learning across NLP tasks}},
    author = {Senge, Manuel and Igamberdiev, Timour and Habernal, Ivan},
    publisher = {Association for Computational Linguistics},
    booktitle = {Proceedings of the 2022 Conference on Empirical
                 Methods in Natural Language Processing},
    pages = {7340--7353},
    year = {2022},
    address = {Abu Dhabi, United Arab Emirates},
    url = {https://arxiv.org/abs/2112.08159}
}
\end{verbatim}
\twocolumn

\maketitle
\begin{abstract}
Preserving privacy in contemporary NLP models allows us to work with sensitive data, but unfortunately comes at a price. We know that stricter privacy guarantees in differentially-private stochastic gradient descent (DP-SGD) generally degrade model performance. However, previous research on the efficiency of DP-SGD in NLP is inconclusive or even counter-intuitive. In this short paper, we provide an extensive analysis of different privacy preserving strategies on seven downstream datasets in five different `typical' NLP tasks with varying complexity using modern neural models based on BERT and XtremeDistil architectures. We show that unlike standard non-private approaches to solving NLP tasks, where bigger is usually better, privacy-preserving strategies do not exhibit a winning pattern, and each task and privacy regime requires a special treatment to achieve adequate performance.
\end{abstract}

\section{Introduction}
\label{sec:intro}

In a world where `data is the new oil', preserving individual privacy is becoming increasingly important. However, modern neural networks are vulnerable to privacy attacks that could even reveal verbatim training data \citep{Carlini.et.al.2020.arXiv}.
An established method for protecting privacy using the \emph{differential privacy} (DP) paradigm \citep{Dwork.Roth.2013} is to train networks with differentially private stochastic gradient descent (DP-SGD) \citep{Abadi.et.al.2016.SIGSAC}. Although DP-SGD has been used in language modeling \citep{McMahan.et.al.2018.ICLR,Hoory.et.al.2021.FindingsEMNLP}, the community lacks a thorough understanding of its usability across different NLP tasks. Some recent observations even seem counter-intuitive, such as the non-decreasing performance at extremely strict privacy values in named entity recognition \citep{Jana.Biemann.2021.PrivNLP}. As such, existing research on the suitability of DP-SGD for various NLP tasks remains largely inconclusive.

We thus ask the following research questions: First, which models and training strategies provide the best trade-off between privacy and performance on different NLP tasks? Second, how exactly do increasing privacy requirements hurt the performance? To answer these questions, we conduct extensive experiments on seven datasets over five tasks, using several contemporary models and varying privacy regimes. Our main contribution is to help the NLP community better understand the various challenges that each task poses to privacy-preserving learning.\footnote{
Code and data at  \url{https://github.com/trusthlt/dp-across-nlp-tasks}}

\section{Related work}

Differential privacy formally guarantees that the probability of leaking information about any individual present in the dataset is proportionally bounded by a pre-defined constant $\varepsilon$, the privacy budget.
We briefly sketch the primary ideas of differential privacy and DP-SGD. For a more detailed introduction please refer to
\citet{Abadi.et.al.2016.SIGSAC,Igamberdiev.Habernal.2022.LREC,Habernal.2021.EMNLP,Habernal.2022.ACL}.

Two datasets are considered \textit{neighboring} if they are identical, apart from one data point (e.g., a document), where each data point is associated with one individual.
A randomized algorithm $\mathcal{M}: \mathcal{X} \rightarrow \mathcal{Y}$ is $(\varepsilon, \delta)$-differentially private if the following probability bound holds true for all neighboring datasets $x, x' \in \mathcal{X}$ and all $y \in \mathcal{Y}$:
\begin{equation}
\Pr \left(
\ln \left[ \frac{\text{Pr}(\mathcal{M}(x) = y)}{\text{Pr}(\mathcal{M}(x') = y)} \right] > \varepsilon
\right) \leq \delta,
\end{equation}

where $\delta$ is a negligibly small constant which provides a relaxation of the stricter $(\varepsilon, 0)$-DP and allows for better composition of multiple differentially private mechanisms (e.g. training a neural model over several epochs). The above guarantee is achieved by adding random noise to the output of $\mathcal{M}$, often drawn from a Laplace or Gaussian distribution. Overall, this process effectively bounds the amount of information that any one individual can contribute to the output of mechanism $\mathcal{M}$.


When using differentially private stochastic gradient descent (DP-SGD) \citep{Abadi.et.al.2016.SIGSAC}, we introduce two additional steps to the standard stochastic gradient descent algorithm. For a given input $x_i$ from the dataset, we obtain the gradient of the loss function $\mathcal{L}(\theta)$ at training time step $t$, $g_t(x_i) = \nabla_{\theta_t} \mathcal{L}(\theta_t, x_i)$. We then clip this gradient by $\ell_2$ norm with clipping threshold $C$ in order to constrain its range, limiting the amount of noise required for providing a differential privacy guarantee.
\begin{equation}
\bar{g}_{t}(x_i) = \frac{g_t (x_i)}{\max \left( 1, \frac{|| g_t(x_i)||_2}{C} \right)}
\end{equation}
Subsequently, we add Gaussian noise to the gradient to make the algorithm differentially private. This DP calculation is grouped into `lots' of size $L$.
\begin{equation}
\displaystyle \tilde{g}_{t} = \frac{1}{L} (\sum_{i \in L} \bar{g}_{t}(x_i) + \mathcal{N}(0, \sigma^2C^2\mathbf{I}))
\end{equation}
The descent step is then performed using this noisy gradient, updating the network's parameters $\theta$, with learning rate $\gamma$.
\begin{equation}
\theta_{t+1} = \theta_t - \gamma \tilde{g}_{t}
\end{equation}


In NLP, several works utilize DP-SGD, primarily for training language models.
\citet{Kerrigan.et.al.2020.PrivNLP} study the effect of using DP-SGD on a GPT-2 model, as well as two simple feed-forward networks, pre-training on a large public dataset and fine-tuning with differential privacy.
Model perplexities are reported on the pre-trained models, but there are no additional experiments on downstream tasks.
\citet{McMahan.et.al.2018.ICLR} train a differentially private LSTM language model that achieves accuracy comparable to non-private models.
\citet{Hoory.et.al.2021.FindingsEMNLP} train a differentially private BERT model and a privacy budget $\varepsilon = 1.1$, achieving comparable accuracy to the non-DP setting on a medical entity extraction task.

Only a few works investigate DP-SGD for downstream tasks in NLP.  \citet{Jana.Biemann.2021.PrivNLP} look into the behavior of differential privacy on the CoNLL 2003 English NER dataset \citep{TjongKimSang.DeMeulder.2003}. They find that no significant drop occurs, even when using low $\varepsilon$ values such as 1, and even as low as 0.022. This is a very unusual result and is assessed in this work further below. \citet{Bagdasaryan.et.al.2019.NeurIPS} apply DP-SGD to sentiment analysis of Tweets, reporting a very small drop in accuracy with epsilons of 8.99 and 3.87. With the state of the art only evaluating on a limited set of tasks and datasets, using disparate privacy budgets and metrics, there is a need for a more general investigation of the DP-SGD framework in the NLP domain, which this paper addresses.

\section{Experimental setup}

\subsection{Tasks and dataset}

We experiment with seven widely-used datasets covering five different standard NLP tasks. These include sentiment analysis (SA) of movie reviews \citep{Maas.et.al.2011} and natural language inference (NLI) \citep{Bowman.et.al.2015} as text classification problems. For sequence tagging, we explore two tasks, in particular named entity recognition (NER) on CoNLL'03 \citep{TjongKimSang.DeMeulder.2003} and Wikiann \citep{Pan.et.al.2017.ACL,Rahimi.et.al.2019.ACL} and part-of-speech tagging (POS) on GUM \citep{Zeldes.2017} and EWT \citep{Silveira.et.al.2014}. The third task type is question answering (QA) on SQuAD~2.0 \citep{Rajpurkar.et.al.2018.ACL}. We chose two sequence tagging tasks, each involving two datasets, to shed light on the surprisingly good results on CoNLL in \citep{Jana.Biemann.2021.PrivNLP}. Table \ref{tab:dataset} summarizes the data statistics.

\begin{table}[ht]
	\resizebox{\linewidth}{!}{
		\begin{tabular}{llrr}
			\toprule
			{Task} & {Dataset} & {Size} & {Classes}\\ 
			\hline
			SA & IMDb & 50k documents & 2\\
			NLI&  SNLI &  570k pairs &  3\vspace{0.7em}\\
			NER &  CoNLL'03 &  $\approx$ 300k tokens &  9 \\
			NER & Wikiann & $\approx$ 320k tokens & 7\\
			POS & GUM & $\approx$ 150k tokens  & 17 \\
			POS & EWT & $\approx$ 254k tokens & 17\vspace{0.7em}\\
			QA&  SQuAD 2.0 & 150k questions  &  $\star$ \\
			\bottomrule
		\end{tabular}
	}
	\caption{Datasets and their specifics. $\star$ SQuAD contains 100k answerable and 50k unanswerable questions, where answerable questions are expressed as the span positions of their answer.}
	\label{tab:dataset}
\end{table}

\subsection{Models and training strategies}
\label{sec:models}

We experiment with five different training (fine-tuning) strategies over two base models. As a simple baseline, we opt for (1) Bi-LSTM to achieve compatibility on NER with previous work  \citep{Jana.Biemann.2021.PrivNLP}.
Further, we employ BERT-base with different fine-tuning approaches. We add (2) LSTM on top of frozen BERT encoder \citep{Fang.et.al.2020.EMNLP} (\texttt{Tr/No/LSTM}), a (3) simple softmax layer on top of `frozen' BERT (\texttt{Tr/No}), and the same configuration with (4) fine-tuning only the last two layers of BERT (\texttt{Tr/Last2}) and finally (5) fine-tuning complete BERT without the input embeddings layer (\texttt{Tr/All}).

In contrast to the non-private setup, the number of trainable parameters affects DP-SGD, since the required noise grows with the gradient size. Therefore we also run the complete experimental setup with a distilled transformer model XtremeDistilTransformer Model (XDTM) \cite{xtreem}. Details of the privacy budget and its computation, as well as hyperparameter tuning are in Appendix \ref{chap:hyper} and \ref{sec:priv}.

\section{Analysis of results}

As LSTM performed worst in all setups, we review only the transformer-based models in detail. Also, we discuss only $\varepsilon$ of $1.0, 5.0,$ and $\infty$ (non-private) as three representative privacy budgets. Here we focus on the BERT-based scenario; a detailed analysis of the distilled model XDTM is provided in Appendix \ref{app:distil}. Random and majority baselines are reported only in the accompanying materials, as they only play a minor role in the drawn conclusions. All results are reported as macro-$F_1$ (averaged $F_1$ of all classes).
We additionally provide an analysis of efficiency and scalability in Appendix \ref{sec:scal} and an analysis on optimal learning rates in Appendix \ref{sec:learning}. 

\begin{figure*}
\includegraphics[width=\linewidth]{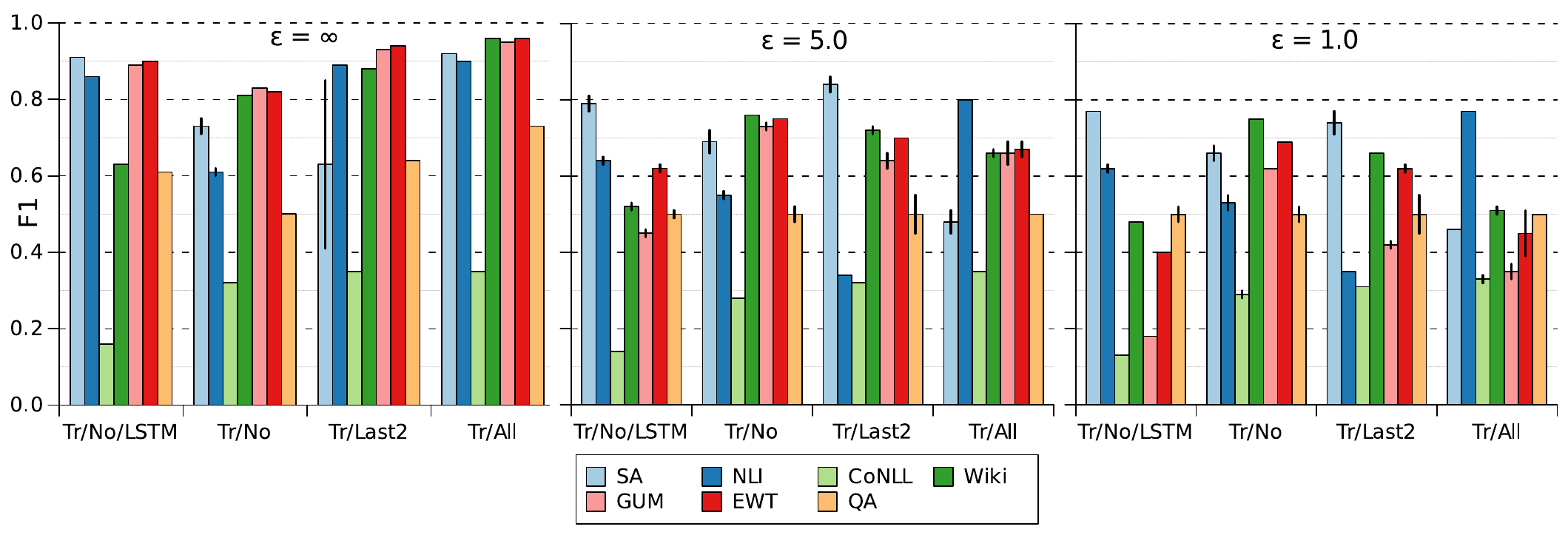}
\caption{\label{fig:results1} Macro $F_1$ scores for non-private ($\varepsilon = \infty$) and two private configurations ($\varepsilon \in \{5; 1\}$) grouped by a particular model (x-axis). Each column represents the score for a specific task performed by the corresponding model. When analyzing one task (one column) in the non-private ($\varepsilon$=$\infty$) setting for different models, macro $F_1$ increases when adding fine-tuning. In DP models, no clear pattern can be observed and the best model is task specific. A complementary task-specific chart with performance drops is shown in Fig.~\ref{fig:performance-by-task} in the Appendix.}
\end{figure*}

\subsection{Sentiment analysis}

Binary sentiment analysis (positive, negative) is a less complex task. Unsurprisingly, all non-private models achieve good results (Figure~\ref{fig:results1} left). Moreover, each model --- except for the fully fine-tuned BERT model \texttt{Tr/All} --- shows only a small performance drop with DP.

\paragraph{Why fine-tuning full BERT with DP-SGD fails?}

While fully fine-tuned BERT is superior in non-private setups, DP-SGD behaves unexpectedly. Instead of having more varied predictions with increasing DP noise, it simply predicts everything as negative with $F_1 = 0.46$, even though the training data is well-balanced (see the confusion matrix in Table~\ref{tab:ConfSA} in the appendix). However, fine-tuning only the last two layers (\texttt{Tr/Last2}) achieves $F_1 > 0.7$. This seems to be consistent with the observation that semantics is spread across the entire BERT model, whereas higher layers are task specific \citep[Sec.~4.3]{Rogers.et.al.2020.BERT}, and that sentiment might be well predicted by local features \citep{Madasu.AnveshRao.2019.EMNLP} which are heavily affected by the noise.

\subsection{Natural Language Inference}

As opposed to sentiment analysis, NLI results show a different pattern. In the non-private setup, two models including fine-tuning BERT, namely \texttt{Tr/Last2} and \texttt{Tr/All}, outperform other models. This is not the case for DP training.

\paragraph{What happened to BERT with DP on the last layers?}

Fine-tuning only the last two layers (\texttt{Tr/Last2}) results in the worst performance in private regimes, e.g., for $\varepsilon = 1$ (Fig.~\ref{fig:results1} right). Analyzing the confusion matrix shows that this model fails to predict neutral entailment, as shown in Table \ref{tab:conf.tr.last2.NLI} top. We hypothesize that the DP noise destroys the higher, task-specific BERT layers, and thus fails on the complexity of the NLI task, that is to recognize cross-sentence linguistic and common-sense understanding. Furthermore, it is easier to recognize entailment or contradiction using some linguistic shortcuts, e.g., similar words pairs, or dataset artifacts \citep{Gururangan.et.al.2018.NAACL.short}.

Full fine-tuning BERT (\texttt{Tr/All}) yields the best private performance, as can be also seen in Table~\ref{tab:conf.tr.last2.NLI} bottom. This shows that noisy gradient training spread across the full model increases robustness for the down-stream task which, unlike sentiment analysis, might not heavily depend on local features, e.g., word n-grams.

\begin{table}[ht]

\begin{center}
\begin{tabular}{rr|rrr}
\toprule
& $\downarrow$ gold & Entail. & Contrad. & Neutral \\
\midrule
\multirow{3}{*}{\rotatebox[origin=c]{90}{ \texttt{Tr/Last2}}} &
Entail. & \textbf{1356} & 1699  & 313 \\
& Contrad. & 1122  & \textbf{1780} & 335 \\
& Neutral & 1217  & 1677  & \textbf{325}\vspace{0.9em}\\
\multirow{3}{*}{\rotatebox[origin=c]{90}{ \texttt{Tr/All}}} &
Entail. & \textbf{2832} & 129   & 407 \\
& Contrad. & 272   & \textbf{2530} & 435 \\
& Neutral & 375   & 604   & \textbf{2240} \\
\bottomrule
\end{tabular}
\end{center}
\vspace{-1em}
\caption{Confusion matrices for $\varepsilon = 1$ on NLI.}
\label{tab:conf.tr.last2.NLI}
\end{table}

\subsection{NER and POS-Tagging}

While the class distribution of SA and NLI is well-balanced, the four datasets chosen for the two sequence tagging tasks are heavily skewed (see Tables \ref{tab:POSData} and \ref{tab:NERData} in the Appendix). Our results show that this imbalance negatively affects all private models.
The degradation of the underrepresented class is known in the DP community. \citet{Farrand.et.al.2020.PPMLP} explore different amounts of imbalances and the effect on differentially private models. \citet{Bagdasaryan.et.al.2019.NeurIPS} find that an imbalanced dataset has only a small negative effect on accuracy for the underrepresented class when training in the non-private setting but it degrades when adding differential privacy.
Both NER and POS-tagging behave similarly when being exposed to DP, as only the most common tags are well predicted, namely the outside tag for NER and the tags for noun, punctuation, verb, pronoun, adposition, and determiner for POS-tagging. Tables \ref{tab:NERF1} and \ref{tab:POSF1pC} in the Appendix show the large differences in $F_1$-scores with $\varepsilon = 1$.

\paragraph{Drawing conclusions using unsuitable metrics?}
While the average of all class $F_1$-scores (macro $F_1$-score) suffers from wrong predictions of underrepresented classes, accuracy remains unaffected. Therefore we suggest using macro $F_1$ to evaluate differential private models which are trained on imbalanced datasets. The difference in accuracy-based metric and macro $F_1$ score explains the unintuitive invariance of NER to DP in \citep{Jana.Biemann.2021.PrivNLP}.

\paragraph{Non-private NER misclassifies prefix but DP fails on tag type.}
Further inspection of the NER results reveals that without DP, the model tends to correctly classify the type of tag (e.g. \textsc{loc}, \textsc{per}, \textsc{org}, \textsc{misc}) but sometimes fails with the position (\textsc{i}, \textsc{b} prefix). This can be seen in the confusion matrix (Table~\ref{tab:NERTrAllnoDP} in the Appendix), examples being \textsc{i-loc} very often falsely classified as \textsc{b-loc} or \textsc{i-per} as \textsc{b-per}. The same pattern is present for \textsc{i-org}, and \textsc{i-misc}. However, DP affects the models' predictions even further, as now, additionally to the position, the tag itself gets wrongly predicted. To exemplify, \textsc{i-misc} is falsely predicted as \textsc{b-loc} 502 times and \textsc{i-org} as \textsc{b-loc} 763 times, as can be seen in Table \ref{tab:NERTrAllDP} in the Appendix.

\paragraph{NER is majority-voting with meaningful $\varepsilon$ already, turns random only with very low $\varepsilon$.}

As we sketched in the introduction, \citet{Jana.Biemann.2021.PrivNLP} showed that a differential private BiLSTM trained on NER shows almost no difference in accuracy compared to the non-private model. However, our experiments show that even with $\varepsilon = 1$, the model almost always predicts the outside tag, as can be seen in the confusion matrix in Table~\ref{tab:NERDP1LSTM} in the Appendix. As mentioned before, the accuracy does not change much since the outside tag is the majority label. Yet, the $F_1$-score more accurately evaluates the models, revealing the misclassifications (CoNLL accuracy: 0.81 vs.\ $F_1$: 0.20; Wikiann accuracy: 0.51 vs.\ $F_1$: 0.1).\footnote{In the non-private setup, this discrepancy remains but it becomes obvious that a more complex model (e.g., BERT) solves the task better.}
Even when choosing much smaller $\varepsilon$, this behavior stays the same. We only were able to get worse results with an extremely low privacy budget $\varepsilon = 0.00837$, which renders the model just predicting randomly (see Table~\ref{tab:smallEps} in the Appendix).

\subsection{Question Answering}

Whereas models with more fine-tuned layers improve non-private predictions (\texttt{Tr/No} $<$ \texttt{Tr/Last2} $<$ \texttt{Tr/All}), with DP, all models drop to 0.5 $F_1$, no matter how strict the privacy is ($\varepsilon = 1, 2, 5$).
We found that throughout all DP models almost all questions are predicted as \emph{unanswerable}. Since 50\% of the test set is labeled as such, this solution allows the model to reach a 0.5 $F_1$ score.\footnote{We used the official evaluation script for SQuAD 2.0} This behavior mirrors the pattern observed in NER and POS-tagging.
Overall, QA is relatively challenging, with many possible output classes in the span prediction process.
For future analyses of DP for QA, we suggest to use Squad v1, as there are no unanswerable questions.

\subsection{Performance drop with stricter privacy}

While a performance drop is unsurprising with decreasing $\varepsilon$, there is no consistent pattern among tasks and models (see Fig.~\ref{fig:performance-by-task} in the Appendix). For instance, while the fully fine-tuned BERT model experiences a relatively large drop for sentiment analysis, its drop for NLI is almost negligible. The actual choice of the model should be therefore taken with a specific privacy requirement in mind.

\section{Conclusion}

We explored differentially-private training on seven NLP datasets. Based on a thorough analysis we formulate three take-home messages.
(1) Skewed class distributions, which are inherent to many NLP tasks, hurt performance with DP-SGD, as the majority classes are often overrepresented in these models. (2) Fine-tuning and thus noisifying different transformer layers affects task-specific behavior so that no single approach generalizes over various tasks, unlike in a typical non-private setup. In other words, there is no one general setup in terms of fine-tuning to achieve the greatest performance for a differentially private model, considering multiple tasks. The best setting is task-specific. (3) Previous works have misinterpreted private NER due to an unsuitable evaluation metric, ignoring class imbalances in the dataset.

\section*{Acknowledgements}

The independent research group TrustHLT is supported by the Hessian Ministry of Higher Education, Research, Science and the Arts. This project was partly supported by the National Research Center for Applied Cybersecurity ATHENE. We thank all the reviewers who tore the paper apart but ultimately helped us make it a much better contribution.

\bibliography{bibliography}
\bibliographystyle{acl_natbib}

\appendix

\section{Hyperparameter tuning}
\label{chap:hyper}
For each model, hyperparametertuning is conducted, and learning rates in the range of \(0.1\) and \(10^{-5}\) are tested. The batch size is set to 32 unless differential privacy and finetuning prohibit such a large size due to memory consumption. If this is the case, the batch size is reduced by a power of 2.\\
For the differentially private models, we obtain the best learning rate for $\varepsilon = 1$ and use it for the same model type with $\varepsilon = 2$ and $\varepsilon = 5$.

\section{Privacy Settings}
\label{sec:priv}
\subsection{Chosen privacy settings}
As the \textit{randomized response} (section 3.2 in \citep{Dwork.Roth.2013}) is considered good privacy with \(\varepsilon = \ln(3) \approx 1.1\), we conduct our experiments with $\varepsilon = 1$. Moreover, as this work aims to understand the behavior of differential privacy, we additionally consider experiments with $\varepsilon = 2$ and $\varepsilon = 5$ to vary the degree of privacy applied to the model. Throughout all our experiments, we set $\delta$ to \(10^{-5}\).

\subsection{Obtaining the privacy parameters}
To achieve a specific (\(\varepsilon\),\(\delta\))-differential privacy guarantee, one has to carefully add the right amount of noise to the gradient during every update step. Unfortunately, it is not possible to calculate the amount of noise needed to achieve a certain (\(\varepsilon\),\(\delta\))-differentially private model in closed-form for DP-SGD. However, the amount of noise only depends on the training dataset size, the number of maximum update steps, as well as $\delta$ and the noise multiplier. Knowing these parameters, one can estimate the resulting $\varepsilon$ value. Exploiting this ability, we iteratively test different sets of parameters and choose the ones that most closely resemble the desired $\varepsilon$.

To calculate the resulting $\varepsilon$ given the parameter set, we use Tensorflow Privacy. In order to incorporate the actual privacy component to the model, we use the Opacus library \citep{opacus}, specifying the amount of noise we calculated for a given $\varepsilon$ value with Tensorflow Privacy.

\section{Efficiency and scalability of differentially private models}
\label{sec:scal}
While the performance of a model is a crucial factor in its evaluation, one other aspect, especially when aiming to reduce energy consumption and development time, is gaining significance. Specifically, this is the time it takes for the model to optimize once across all batches and then evaluate itself against the evaluation dataset (epoch time).

A significant increase in epoch time is observed in all experiments when training in the differentially private setting. This increase is possibly related to the number of entries in the dataset. As shown in Table \ref{tab:timingInc}, larger datasets such as SNLI (570k entries) for NLI or SQuAD 2.0 (150k entries) for QA present a significantly larger increase in epoch time, compared to tasks trained on smaller datasets (such as IMDb for SA (50k entries), CoNLL’03 for NER (15k entries) or EWT for POS-tagging (13k entries)).

Another possible influence on the epoch time is number of fine-tuned parameters of the model. When adding further fine-tuning, the increase in epoch time tends to get larger. Throughout all tasks shown in Table \ref{tab:timingInc}, a larger difference is shown for the fully fine-tuned transformer \texttt{(Tr/All)}, as when only fine-tuning the last two layers \texttt{(Tr/Last2)}.
However, future work further examining these possible correlations is necessary for providing a clearer picture.

\begin{table*}[ht]
\begin{center}\resizebox{\textwidth}{!}{%

    \begin{tabular}{c c c c c}
    \toprule
    \multicolumn{5}{c}{Epoch time differences with and without differential privacy (DP)} \\
    \midrule
    Task  & \multicolumn{1}{p{7.785em}}{fine-tuned Layers\newline{}of BERT base} & epoch time without DP & epoch time with DP &\cellcolor[rgb]{ .949,  .949,  .949} difference \\
    \midrule
    SA & \texttt{Tr/Last2} & 4 min 22 sec & 0 h 10 min 32 sec &\cellcolor[rgb]{ .949,  .949,  .949} 0 h 06 min 10 sec \\
    SA & \texttt{Tr/All}   & 9 min 07 sec & 0 h 47 min 08 sec &\cellcolor[rgb]{ .949,  .949,  .949} 0 h 38 min 01 sec \\
    \midrule
    NER  (CoNLL’03) & \texttt{Tr/Last2} & 0 min 26 sec & 0 h 02 min 24 sec &\cellcolor[rgb]{ .949,  .949,  .949} 0 h 01 min 58 sec \\
    NER  (CoNLL’03) & \texttt{Tr/All}  & 0 min 57 sec & 0 h 27 min 55 sec &\cellcolor[rgb]{ .949,  .949,  .949} 0 h 26 min 58 sec \\
    \midrule
    NLI   & \texttt{Tr/Last2} & 13 min 15 sec & 3 h 37 min 15 sec &\cellcolor[rgb]{ .949,  .949,  .949} 3 h 24 min 00 sec \\
    NLI   & \texttt{Tr/All}   & 22 min 32 sec & 9 h 57 min 57 sec &\cellcolor[rgb]{ .949,  .949,  .949} 9 h 35 min 25 sec \\
    \midrule
    POS-tagging (EWT) & \texttt{Tr/Last2} & 0 min 53 sec & 0 h 06 min 12 sec &\cellcolor[rgb]{ .949,  .949,  .949} 0 h 05 min 19 sec \\
    POS-tagging (EWT) & \texttt{Tr/All}  & 1 min 12 sec & 0 h 49 min 25 sec &\cellcolor[rgb]{ .949,  .949,  .949} 0 h 48 min 13 sec \\
    \midrule
    QA    & \texttt{Tr/Last2} & 12 min 21 sec & 1 h 57 min 43 sec &\cellcolor[rgb]{ .949,  .949,  .949} 1 h 45 min 22 sec \\
    QA    & \texttt{Tr/All}  & 44 min 07 sec & 11 h 05 min 15 sec &\cellcolor[rgb]{ .949,  .949,  .949} 10 h 21 min 08 sec \\
    \bottomrule
    \end{tabular}%

}\end{center}

\caption[Epoch time for each task and its difference for the DP and no DP setting]{Average epoch time during the training process for each task and two transformer fine-tuning settings (\texttt{Tr/All} and \texttt{Tr/Last2}). Right-most column shows the difference between the differentially private and non-private models.}
\label{tab:timingInc}
\end{table*}

\section{Analysis of the learning rate}
\label{sec:learning}

For non-private models, using smaller learning rates usually leads to slower convergence towards a local minimum. This effect can be compensated when training for enough epochs. When considering a training process using differential privacy, this assumption no longer holds. One reason is that with every update step, random noise is added to the model, or in other words, no step is exactly towards the optimal direction. Thus, a smaller learning rate, or more necessary steps to reach a local optimum, increases the probability in moving in the wrong direction.

\begin{figure}
  \centering \includegraphics[width=\linewidth]{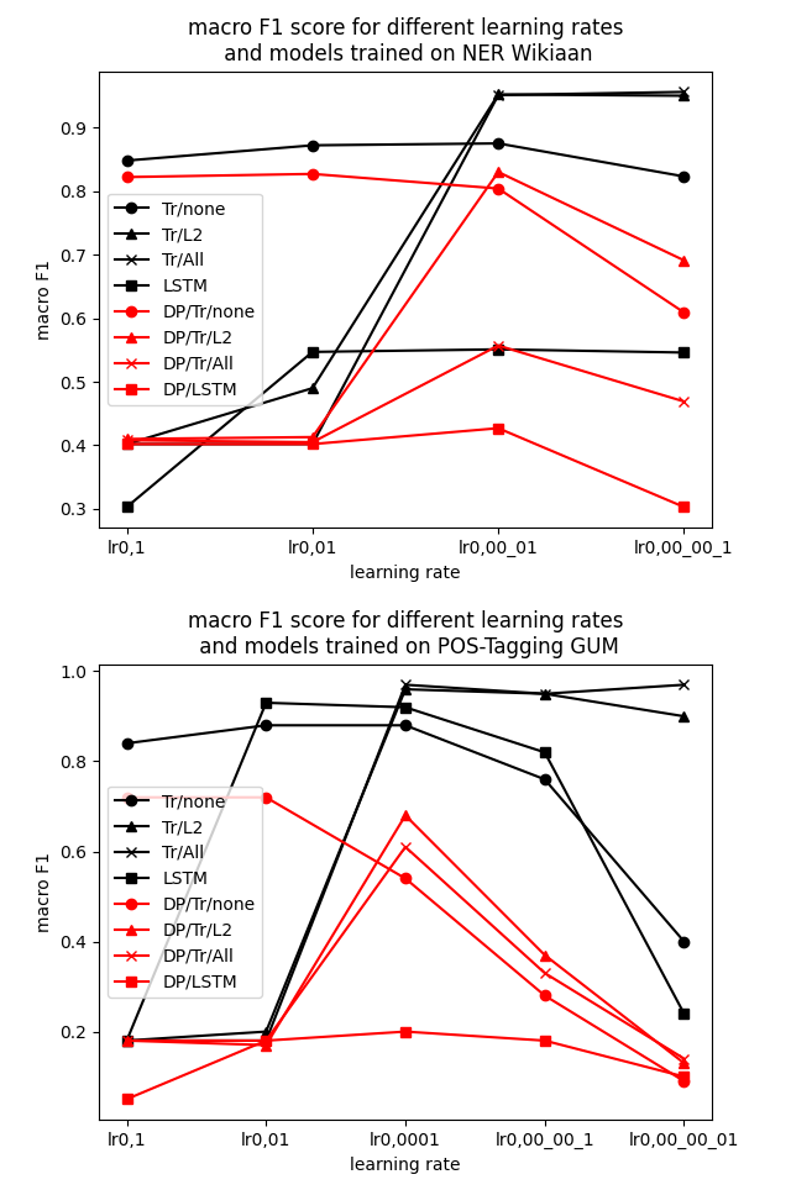}
  \caption{Both plots display the different macro $F_1$ scores for each model and the corresponding learning rate. It is notable, that for the differentially private models a smaller learning rate tends to (more quickly) result in worse performance.}
  \label{fig:learning}
\end{figure}

This behavior can be observed in our experiments as well. When looking at Figure \ref{fig:learning}, the macro $F_1$ for the non-private experiments (black) usually increases or converges when decreasing the learning rate. However, when analyzing the experiments with DP (red) it is notable that with a smaller learning rate, the macro $F_1$ score decreases earlier than in the setting without DP.

\section{Analysing XtremeDistilTransformer Model}
\label{app:distil}
To compare our results and achieve a more robust analysis, we repeated our experiments using the same setup as before, and trained the XtremeDistilTransformer Model (XDTM) \cite{xtreem}.

\subsection{Comparing both non-private models}
When comparing the XDTM with BERT and no differential privacy, we achieve similar behavior. It is however notable, that for SA the \texttt{XDTM/Last2} setup is much better than \texttt{Tr/Last2}. Similar results can be seen for NLI where only \texttt{(Tr/No)} presents better results compared to the BERT model. Furthermore, the XDTM trained on CoNLL fails to learn the task as it not only mistakes the I and B prefix, but also the tag itself. However, when training the XDTM on Wikiann, we can see much better performance. One reason could be the reduced tag size (9 for CoNLL and 7 for Wikiann).
When looking at POS-tagging, we can see similar behavior for both GUM and EWT. Here the best choice is either the \texttt{XDTM/No/LSTM} or \texttt{XDTM/All}. Both \texttt{XDTM/Last2} or \texttt{XDTM/No} show significantly lower performance. This behavior comes as a surprise, as the BERT model performs similarly in every setting.
For QA, the non-private models are all less accurate except for XDTM with an additional LSTM.

\subsection{Differentially Private XtremeDistilTransformer Model}
When analyzing the different tasks and how the models behave when introduced to differential privacy, we can group them into two categories.
While SA, NLI barely show a drop in macro $F_1$, NER on Wikiann as well as POS-tagging on GUM and EWT display a larger drop (Wikiann: between 14\% and 4\%, GUM: between 73\% and 19\%, EWT: between 66\% and 19\%). Additionally, as the non private model trained on CoNLL already had very low performance, it can be ignored in this analysis.
We could not run QA experiments because of a limited access to further GPU compute capacity.
For QA, the fully finetuned XDTM almost always predicts `unanswerable'. In contrast, the XDTM with an additional LSTM does predict spans. These predictions however, result in a worse performance of the model (about 0.33 $F_1$), compared to a model only predicting unanswerable (about 0.5 $F_1$).

\subsection{Compare the differentially private setting to BERT}
The main difference in differentially private SA between BERT and XDTM is that XDTM has no significant drop for all settings. Contrarily, BERT shows a large performance decline when fine-tuning all layers.

When training XDTM on NLI, it shows a better performance than the BERT model. This comes to a surprise, as one would expect XDTM to have a lower accuracy, since it was trained to mimic BERT's behavior and has less parameters.

For NER with CoNLL, on the one hand both models show a bad macro $F_1$, where XDTM is about 15\% worse than BERT. On the other hand, when training the two models on Wikiann, we can see that they are about the same if we include an LSTM after the (XtremeDistil) BERT layer. When omitting this extra network we can see a significant drop (about 40\%) for XDTM.

For both POS-tagging tasks, XDTM is significantly worse than BERT, between 29\% and 14\% for GUM and 40\% to 14\% for EWT.
For QA, the performance of the differentially-private XDTM is about the same as for BERT. However, the additional LSTM seems to be worse for the XDTM as for BERT. The reason is that XDTM predicts a lot of false spans. This is not the case for \texttt{Tr/No/LSTM}, as it only predicts `unanswerable'.

\subsection{Conclusion for differential private XtremeDistilTransformer Model}
To conclude, our experiments show that using XDTM for the tasks with differential privacy can be useful for simple classification tasks with small numbers of classes (such as SA or NLI). However, if the number of classes increase (such as NER or POS-tagging), differentially private XDTMs tend to perform worse than BERT. It is worth mentioning, however, that training differentially private XDTMs is much faster than using the full BERT model. See table \ref{tab:timing_xtreem} for the exact difference. Additionally, this speedup is also a direct result of larger possible batch sizes. This is made possible as the memory needed for these models is less than for the full BERT model.

\begin{table*}[ht]
	\begin{center}\resizebox{\linewidth}{!}{%

    \begin{tabular}{c c | c c c|c c c}
    \toprule
    \multicolumn{8}{c}{Runtime differences between the BERT model and the XtremeDistilTransformer Model} \\
    \midrule
    Task  & \multicolumn{1}{p{7.785em}|}{Finetuned Layers\newline{}of BERT Model} & \multicolumn{1}{p{9em}}{BERT\newline{}runtime without DP} & \multicolumn{1}{p{9em}}{XDTM\newline{}runtime without DP} & \cellcolor[rgb]{ .851,  .851,  .851}difference no DP & \multicolumn{1}{p{8em}}{BERT\newline{}runtime with DP} & \multicolumn{1}{p{7.43em}}{XDTM\newline{}runtime with DP} & \cellcolor[rgb]{ .851,  .851,  .851}difference DP \\
    \midrule
    SA & last 2 & 4 min 22 sec & 0 min 21 sec & \cellcolor[rgb]{ .851,  .851,  .851}4 min 01 sec & 0 h 10 min 32 sec & 0 min 21 sec & \cellcolor[rgb]{ .851,  .851,  .851}0 h 10 min 11 sec \\
    SA & all   & 9 min 07 sec & 0 min 52 sec & \cellcolor[rgb]{ .851,  .851,  .851}8 min 15 sec & 0 h 47 min 08 sec & 14 min 29 sec & \cellcolor[rgb]{ .851,  .851,  .851}0 h 32 min 39 sec \\
    \midrule
    NER   & last 2 & 0 min 26 sec & 0 min 16 sec & \cellcolor[rgb]{ .851,  .851,  .851}0 min 10 sec & 0 h 02 min 24 sec & 0 min 17 sec & \cellcolor[rgb]{ .851,  .851,  .851}0 h 2 min 07 sec \\
    NER   & all   & 0 min 57 sec & 0 min 25 sec & \cellcolor[rgb]{ .851,  .851,  .851}0 min 32 sec & 0 h 27 min 55 sec & 1 min 34 sec & \cellcolor[rgb]{ .851,  .851,  .851}0 h 26 min 21 sec  \\
    \midrule
    NLI   & last 2 & 13 min 15 sec & 2 min 34 sec & \cellcolor[rgb]{ .851,  .851,  .851}10 min 41 sec & 3 h 37 min 15 sec & 3 min 36 sec & \cellcolor[rgb]{ .851,  .851,  .851}3 h 33min 39 sec \\
    NLI   & all   & 22 min 32 sec & 18 min 32 sec & \cellcolor[rgb]{ .851,  .851,  .851}4 min 00 sec & 9 h 57 min 57 sec & 62 min 32 sec & \cellcolor[rgb]{ .851,  .851,  .851}8 h 55 min 25 sec  \\
    \midrule
    EWT & last 2 & 0 min 53 sec & 0 min 16 sec & \cellcolor[rgb]{ .851,  .851,  .851}0 min 37 sec & 0 h 06 min 12 sec & 0 min 15 sec & \cellcolor[rgb]{ .851,  .851,  .851}0 h 06 min 03 sec \\
    EWT & all   & 1 min 12 sec & 0 min 19 sec & \cellcolor[rgb]{ .851,  .851,  .851}0 min 53 sec & 0 h 49 min 25 sec & 1 min 32 sec & \cellcolor[rgb]{ .851,  .851,  .851}0 h 47 min 53 sec \\
    \midrule
    QA    & last 2 & 12 min 21 sec & 6 min 45 sec & \cellcolor[rgb]{ .851,  .851,  .851}5 min 36 sec & 1 h 57 min 43 sec & 6 min 32 sec & \cellcolor[rgb]{ .851,  .851,  .851}1 h 51 min 11 sec \\
    QA    & all   & 44 min 07 sec & 7 min 24 sec & 36 min 43 sec \cellcolor[rgb]{ .851,  .851,  .851} & 11 h 05 min 15 sec & 53 min 35 sec & 10 h 11 min 40 sec \cellcolor[rgb]{ .851,  .851,  .851}
    \\
    \midrule
    \end{tabular}%

	}\end{center}

	\caption{The difference in epoch time for the BERT and XDTM model; The time is the average epoch time during the training process. It can be seen, that the XDTM needs a lot less time to train. There are two main reasons for this expected behavior. (1) The XDTM is smaller, therefore, fewer parameters need to be trained. (2) The smaller model uses less memory, therefore, a larger batch size can be used.}
	\label{tab:timing_xtreem}
\end{table*}

\section{Detailed tables and figures}

\begin{figure*}
	\includegraphics[width=\linewidth]{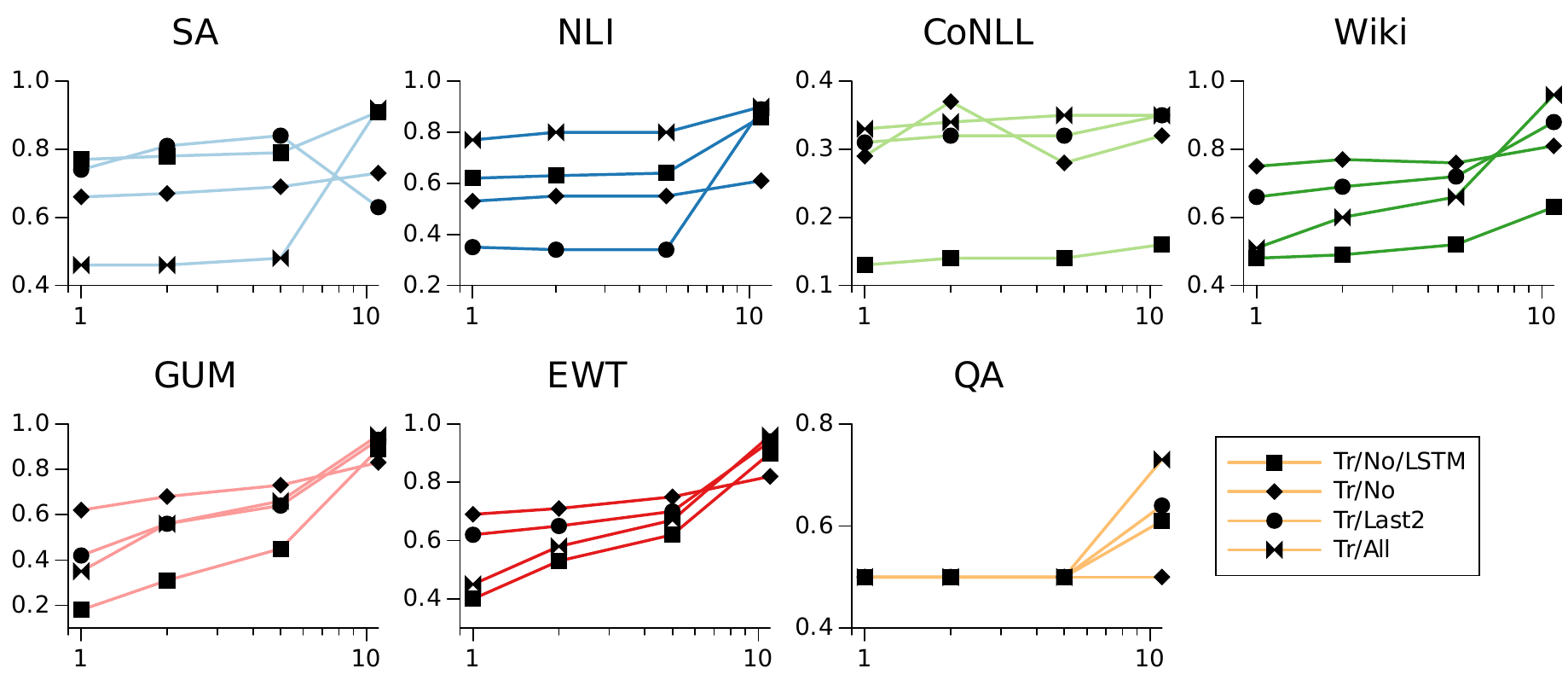}
	\caption{\label{fig:performance-by-task} Comparison of BERT performances (macro $F_1$ score) per dataset with varying privacy budget $\varepsilon \in \{1, 2, 5, \infty \}$ on the $x$-axis (note the $\log$ scale).}
\end{figure*}

\begin{figure*}
	\includegraphics[width=\linewidth]{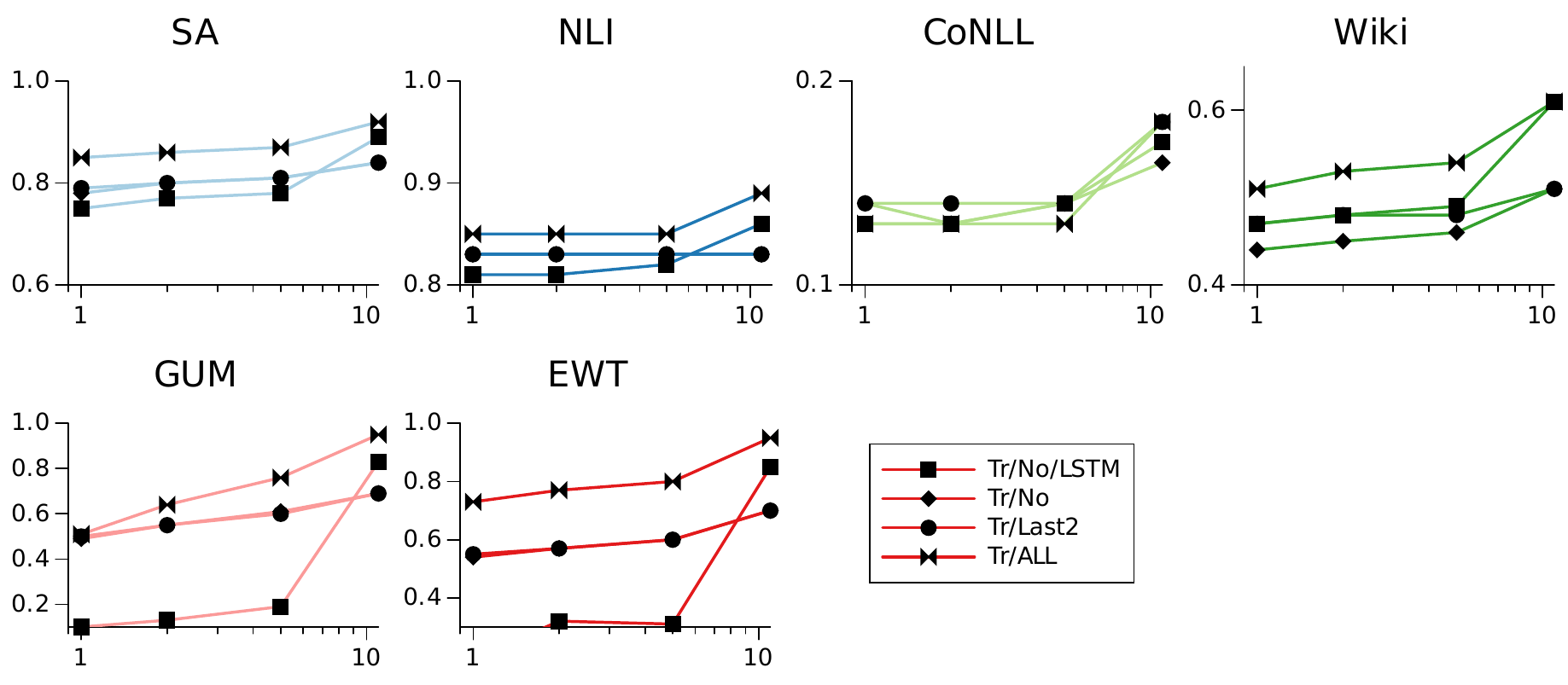}
	\caption{\label{fig:performance-by-task-distil} Comparison of XDTM performances (macro $F_1$ score) per dataset with varying privacy budget $\varepsilon \in \{1, 2, 5, \infty \}$ on the $x$-axis (note the $\log$ scale).}
\end{figure*}

\begin{figure*}
	\includegraphics[width=\linewidth]{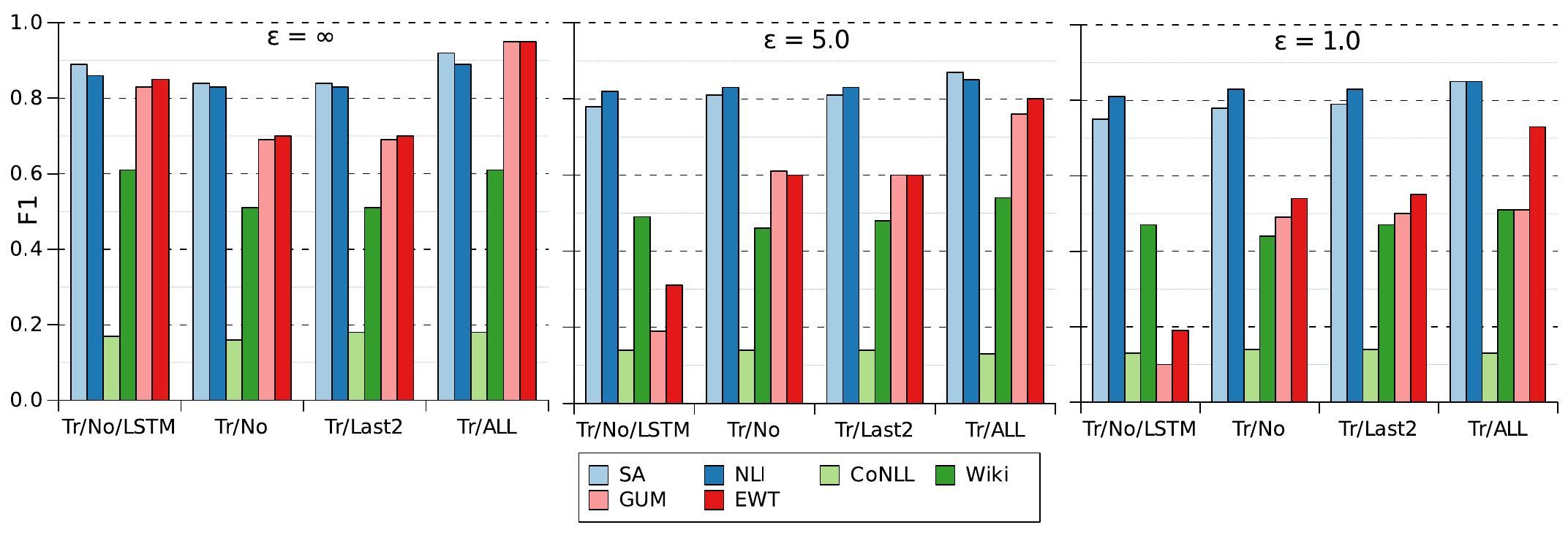}
	\caption{\label{fig:results-distil} Macro $F_1$ scores for non-private ($\varepsilon = \infty$) and two private configurations ($\varepsilon \in \{5; 1\}$) grouped by a particular model (x-axis) with XDTM as a base model.}
\end{figure*}

\begin{table}[ht]
	\centering
	\begin{tabular}{r|rr}
		\toprule
		$\downarrow$ gold & Neg & Pos \\
		\midrule
		Neg & \textbf{12497} & 3 \\
		Pos & 12497 & \textbf{2} \\
		\bottomrule
	\end{tabular}
	\caption{Confusion matrix for the fully fine-tuned BERT model \texttt{Tr/All} with $\varepsilon = 1$ on sentiment analysis.}
	\label{tab:ConfSA}
\end{table}

\begin{table}[ht]
	\centering
	\begin{tabular}{r|rr}
		\toprule
		$\downarrow$ gold & Neg & Pos \\
		\midrule
		Neg & \textbf{21137} & 3863 \\
		Pos & 3162 & \textbf{21838} \\
		\bottomrule
	\end{tabular}
	\caption{Confusion matrix for \texttt{Tr/None/LSTM} with $\varepsilon = 1$ on sentiment analysis.}
	\label{tab:ConfSATrNoneLSTM}
\end{table}

\begin{table}[ht]
	\begin{center}\resizebox{\linewidth}{!}{%

			\begin{tabular}{l|rr|rr}
			\toprule
				& \multicolumn{2}{c|}{{GUM}} & \multicolumn{2}{c}{{EWT}} \\
				
				 & {Train} & {Test} & {Train} & {Test} \\
				\midrule
				{NOUN} & 17,873 & 2,942  & 34,781 & 4,132 \\
				{PUNCT} & 13,650 & 1,985  & 23,679 & 3,106 \\
				{VERB} & 10,957 & 1,647  & 23,081 & 2,655 \\
				{PRON} & 7,597  & 1,128  & 18,577 & 2,158 \\
				{ADP} & 10,237 & 1,698  & 17,638 & 2,018 \\
				{DET} & 8,334  & 1,347  & 16,285 & 1,896 \\
				{PROPN} & 7,066  & 1,230  & 12,946 & 2,076 \\
				{ADJ} & 6,974  & 1,116  & 12,477 & 1,693 \\
				{AUX} & 4,791  & 719   & 12,343 & 1,495 \\
				{ADV} & 4,180  & 602   & 10,548 & 1,225 \\
				{CCONJ} & 3,247  & 587   & 6,707  & 739 \\
				{PART} & 2,369  & 335   & 5,567  & 630 \\
				{NUM} & 2,096  & 333   & 3,999  & 536 \\
				{SCONJ} & 2,095  & 251   & 3,843  & 387 \\
				{X} & 244   & 24    & 847   & 139 \\
				{INTJ} & 392   & 87    & 688   & 120 \\
				{SYM} & 156   & 35    & 599   & 92 \\
				\bottomrule
			\end{tabular}%

	}\end{center}

	\caption{Distribution of all possible tags for POS-Tagging}
	\label{tab:POSData}
\end{table}

\begin{table}[ht]
	\begin{center}\resizebox{\linewidth}{!}{%

			\begin{tabular}{l|rr|rr}
			\toprule
				& \multicolumn{2}{c|}{{CoNLL’03 }} & \multicolumn{2}{c}{{Wikiann}} \\

				 & {Test} & {Train} & {Test} & {Train} \\
				\midrule
				{O} & 38,554 & 170,524 & 42,879 & 85,665 \\
				{I-LOC} & 1,919  & 1,157  & 6,661  & 13,664 \\
				{B-PER} & 0     & 6,600  & 4,649  & 9,345 \\
				{I-PER} & 2,773  & 4,528  & 7,721  & 15,085 \\
				{I-ORG} & 2,491  & 3,704  & 11,825 & 23,668 \\
				{I-MISC} & 909   & 1,155  & --     & -- \\
				{B-MISC} & 9     & 3,438  & --     & -- \\
				{B-LOC} & 6     & 7,140  & 5,023  & 10,081 \\
				{B-ORG} & 5     & 6,321  & 4,974  & 9,910 \\
				\bottomrule
			\end{tabular}%

	}\end{center}

	\caption{Distribution of all possible tags for NER}
	\label{tab:NERData}
\end{table}

\begin{table}[ht]
	\begin{center}
		\begin{tabular}{l|r|r}
			\toprule
			& CoNLL'03 & Wikiann \\
			\midrule
			\rowcolor[rgb]{ .851,  .851,  .851} {O} & 0.98  & 0.86 \\
			{I-LOC} & 0.00  & 0.46 \\
			{B-PER} & 0.00  & 0.69 \\
			{I-PER} & \cellcolor[rgb]{ .851,  .851,  .851} 0.67  & 0.57 \\
			{I-ORG} & 0.01  & 0.54 \\
			{I-MISC} & 0.00  & -- \\
			{B-MISC} & 0.00  & -- \\
			{B-LOC} & 0.00  & 0.44 \\
			{B-ORG} & 0.00  & 0.14 \\
			\bottomrule
		\end{tabular}
	\end{center}
	
	\caption{F1-scores per class for NER using the fully fine-tuned BERT \texttt{Tr/All} and $\varepsilon = 1$. Except for the \textbf{O} tag (highlighted) and \textbf{I-PER}, no other class is predicted with sufficient performance on CoNLL.}
	\label{tab:NERF1}
\end{table}

\begin{table}[ht]
	\begin{center}
		\begin{tabular}{l|rr}
		\toprule
			& GUM & EWT \\
			\midrule
			{NOUN} & \cellcolor[rgb]{ .851,  .851,  .851}0.66 & \cellcolor[rgb]{ .851,  .851,  .851}0.62 \\
			{PUNCT} & \cellcolor[rgb]{ .851,  .851,  .851}0.85 & \cellcolor[rgb]{ .851,  .851,  .851}0.87 \\
			{VERB} & \cellcolor[rgb]{ .851,  .851,  .851}0.64 & \cellcolor[rgb]{ .851,  .851,  .851}0.72 \\
			{PRON} & \cellcolor[rgb]{ .851,  .851,  .851}0.65 & \cellcolor[rgb]{ .851,  .851,  .851}0.72 \\
			{ADP} & \cellcolor[rgb]{ .851,  .851,  .851}0.73 & \cellcolor[rgb]{ .851,  .851,  .851}0.80 \\
			{DET} & \cellcolor[rgb]{ .851,  .851,  .851}0.81 & \cellcolor[rgb]{ .851,  .851,  .851}0.83 \\
			{PROPN} & 0.17  & 0.16 \\
			{ADJ} & 0.13  & 0.03 \\
			{AUX} & 0.41  & 0.69 \\
			{ADV} & 0.00  & 0.10 \\
			{CCONJ} & 0.06  & 0.02 \\
			{PART} & 0.00  & 0.00 \\
			{NUM} & 0.00  & 0.00 \\
			{SCONJ} & 0.00  & 0.00 \\
			{X} & 0.00  & 0.00 \\
			{INTJ} & 0.00  & 0.00 \\
			{SYM} & 0.00  & 0.00 \\
			\bottomrule
		\end{tabular}
	\end{center}

	\caption{F1-scores per class for POS-tagging using the fully fine-tuned BERT \texttt{Tr/All} and $\varepsilon = 1$. Highlighted rows achieve usable performance.}
	\label{tab:POSF1pC}
\end{table}

\begin{table*}[ht]
	\begin{center}
		\begin{tabular}{l|rrrrrrrrr}
			\toprule
			$\downarrow$ gold & O & B-PER & I-PER & B-ORG & I-ORG & B-LOC & I-LOC & B-MISC & I-MISC \\
			\midrule
			O & \textbf{38207} & 26 & 3 & 47 & 41 & 19 & 9 & 66 & 80 \\
			B-PER & 0 & \textbf{0} & 0 & 0 & 0 & 0 & 0 & 0 & 0 \\
			I-PER & 17 & \cellcolor[rgb]{ .851,  .851,  .851}1556 & \cellcolor[rgb]{ .851,  .851,  .851}\textbf{1150} & 23 & 10 & 15 & 1 & 1 & 0 \\
			B-ORG & 0 & 0 & 0 & \textbf{0} & 5 & 0 & 0 & 0 & 0 \\
			I-ORG & 42 & 27 & 3 & \cellcolor[rgb]{ .851,  .851,  .851}1514 & \cellcolor[rgb]{ .851,  .851,  .851}\textbf{767} & 53 & 29 & 42 & 14 \\
			B-LOC & 0 & 0 & 1 & 0 & 2 & \textbf{0} & 1 & 0 & 2 \\
			I-LOC & 22 & 7 & 3 & 49 & 12 & \cellcolor[rgb]{ .851,  .851,  .851}1558 & \cellcolor[rgb]{ .851,  .851,  .851}\textbf{232} & 30 & 5 \\
			B-MISC & 0 & 0 & 0 & 0 & 0 & 0 & 0 & \cellcolor[rgb]{ .851,  .851,  .851}\textbf{4} & \cellcolor[rgb]{ .851,  .851,  .851}5 \\
			I-MISC & 51 & 9 & 3  & 32  & 9  & 23  & 4  & \cellcolor[rgb]{ .851,  .851,  .851}594  & \cellcolor[rgb]{ .851,  .851,  .851}\textbf{183} \\
			\bottomrule
		\end{tabular}%
	\end{center}

	\caption{Confusion matrix for NER on CoNLL'03 for the fully fine-tuned BERT model. It can be seen (highlighted) that the model sometimes falsely predicts the position of the tag. Yet, the tag itself is mostly correctly classified.}
	\label{tab:NERTrAllnoDP}
\end{table*}

\begin{table*}[ht]
	\begin{center}

		\begin{tabular}{l|rrrrrrrrr}
			\toprule
			$\downarrow$ gold & O & B-PER & I-PER & B-ORG & I-ORG & B-LOC & I-LOC & B-MISC & I-MISC \\
			\midrule
			O & \textbf{37540} & 61 & 38 & 149 & 0 & 304 & 0 & 0 & 0 \\
			B-PER & 0 & \textbf{0} & 0 & 0 & 0 & 0 & 0 & 0 & 0 \\
			I-PER & 33 & 1057 & \textbf{1521} & 77 & 0 & 48 & 0 & 0 & 0 \\
			B-ORG & 0 & 0 & 0 & \textbf{4} & 0 & 1 & 0 & 0 & 0 \\
			I-ORG & 192 & 90 & 142 & \cellcolor[rgb]{ .851,  .851,  .851}1157 & \cellcolor[rgb]{ .851,  .851,  .851}\textbf{9} & 763 & 0 & 0 & 0 \\
			B-LOC & 0 & 1 & 2 & 2 & 0 & \textbf{1} & 0 & 0 & 0 \\
			I-LOC & 58 & 14 & 38 & 169 & 0 & 1577 & \textbf{0} & 0 & 0 \\
			B-MISC & 1 & 2 & 0 & 1 & 0 & 5 & 0 & \textbf{0} & 0 \\
			I-MISC  & 178  & 21  & 14  & 110  & 0  & \cellcolor[rgb]{ .851,  .851,  .851}502  & 0  & 0  & \cellcolor[rgb]{ .851,  .851,  .851}\textbf{0} \\
			\bottomrule
		\end{tabular}
	\end{center}

	\caption{Confusion matrix for NER on CoNLL'03 for the fully fine-tuned BERT model with DP ($\varepsilon = 1$). It can be seen (highlighted) that the model sometimes falsely predicts the type of tag. The positional error is still present.}
	\label{tab:NERTrAllDP}
\end{table*}

\begin{table*}[ht]
	\begin{center}
		\begin{tabular}{l|rrrrrrrrr}
			\toprule
			$\downarrow$ gold & O & I-LOC & B-PER & I-PER & I-ORG & I-MISC & B-MISC & B-LOC & B-ORG \\
			\midrule
			O & \cellcolor[rgb]{ .851,  .851,  .851}\textbf{41021} & 52    & 9     & 6     & 9     & 32    & 4     & 4     & 5 \\
			I-LOC & \cellcolor[rgb]{ .851,  .851,  .851}1935 & \textbf{0} & 1     & 0     & 0     & 1     & 0     & 0     & 1 \\
			B-PER & \cellcolor[rgb]{ .851,  .851,  .851}0 & 0     & \textbf{0} & 0     & 0     & 0     & 0     & 0     & 0 \\
			I-PER & \cellcolor[rgb]{ .851,  .851,  .851}2821 & 2     & 1     & \textbf{0} & 0     & 3     & 0     & 0     & 0 \\
			I-ORG & \cellcolor[rgb]{ .851,  .851,  .851}2524 & 3     & 1     & 1     & \textbf{0} & 2     & 1     & 0     & 0 \\
			I-MISC & \cellcolor[rgb]{ .851,  .851,  .851}1013 & 1     & 0     & 0     & 0     & \textbf{1} & 0     & 0     & 0 \\
			B-MISC & \cellcolor[rgb]{ .851,  .851,  .851}9 & 0     & 0     & 0     & 0     & 0     & \textbf{0} & 0     & 0 \\
			B-LOC & \cellcolor[rgb]{ .851,  .851,  .851}6 & 0     & 0     & 0     & 0     & 0     & 0     & \textbf{0} & 0 \\
			B-ORG & \cellcolor[rgb]{ .851,  .851,  .851}5 & 0     & 0     & 0     & 0     & 0     & 0     & 0     & \textbf{0} \\
			\bottomrule
		\end{tabular}
	\end{center}

	\caption{Confusion matrix for NER on CoNLL'03 for the BILSTM with DP ($\varepsilon = 1$). It can be seen that this model only predicts the outside tag (highlighted).}
	\label{tab:NERDP1LSTM}
\end{table*}

\begin{table*}[ht]
	\begin{center}
		\begin{tabular}{l|rrrrrrrrr}
			\toprule
			$\downarrow$ gold & O & I-LOC & B-PER & I-PER & I-ORG & I-MISC & B-MISC & B-LOC & B-ORG \\
			\midrule
			O & \cellcolor[rgb]{ .851,  .851,  .851}\textbf{3464} & \cellcolor[rgb]{ .851,  .851,  .851}3190 & \cellcolor[rgb]{ .851,  .851,  .851}4161 & \cellcolor[rgb]{ .851,  .851,  .851}21208 & \cellcolor[rgb]{ .851,  .851,  .851}170 & \cellcolor[rgb]{ .851,  .851,  .851}6035 & \cellcolor[rgb]{ .851,  .851,  .851}722 & \cellcolor[rgb]{ .851,  .851,  .851}1993 & \cellcolor[rgb]{ .851,  .851,  .851}144 \\
			I-LOC & \cellcolor[rgb]{ .851,  .851,  .851}130 & \cellcolor[rgb]{ .851,  .851,  .851}\textbf{153} & \cellcolor[rgb]{ .851,  .851,  .851}252 & \cellcolor[rgb]{ .851,  .851,  .851}1032 & \cellcolor[rgb]{ .851,  .851,  .851}11 & \cellcolor[rgb]{ .851,  .851,  .851}243 & \cellcolor[rgb]{ .851,  .851,  .851}29 & \cellcolor[rgb]{ .851,  .851,  .851}84 & \cellcolor[rgb]{ .851,  .851,  .851}2 \\
			B-PER & 0     & 0     & \textbf{0} & 0     & 0     & 0     & 0     & 0     & 0 \\
			I-PER & \cellcolor[rgb]{ .851,  .851,  .851}234 & \cellcolor[rgb]{ .851,  .851,  .851}226 & \cellcolor[rgb]{ .851,  .851,  .851}348 & \cellcolor[rgb]{ .851,  .851,  .851}\textbf{1440} & \cellcolor[rgb]{ .851,  .851,  .851}20 & \cellcolor[rgb]{ .851,  .851,  .851}377 & \cellcolor[rgb]{ .851,  .851,  .851}34 & \cellcolor[rgb]{ .851,  .851,  .851}140 & \cellcolor[rgb]{ .851,  .851,  .851}3 \\
			I-ORG & \cellcolor[rgb]{ .851,  .851,  .851}271 & \cellcolor[rgb]{ .851,  .851,  .851}184 & \cellcolor[rgb]{ .851,  .851,  .851}310 & \cellcolor[rgb]{ .851,  .851,  .851}1161 & \cellcolor[rgb]{ .851,  .851,  .851}\textbf{9} & \cellcolor[rgb]{ .851,  .851,  .851}430 & \cellcolor[rgb]{ .851,  .851,  .851}39 & \cellcolor[rgb]{ .851,  .851,  .851}115 & \cellcolor[rgb]{ .851,  .851,  .851}8 \\
			I-MISC & \cellcolor[rgb]{ .851,  .851,  .851}76 & \cellcolor[rgb]{ .851,  .851,  .851}73 & \cellcolor[rgb]{ .851,  .851,  .851}100 & \cellcolor[rgb]{ .851,  .851,  .851}561 & \cellcolor[rgb]{ .851,  .851,  .851}4 & \cellcolor[rgb]{ .851,  .851,  .851}\textbf{132} & \cellcolor[rgb]{ .851,  .851,  .851}17 & \cellcolor[rgb]{ .851,  .851,  .851}49 & \cellcolor[rgb]{ .851,  .851,  .851}3 \\
			B-MISC & 1     & 0     & 1     & 4     & 0     & 0     & \textbf{0} & 3     & 0 \\
			B-LOC & 0     & 1     & 1     & 4     & 0     & 0     & 0     & \textbf{0} & 0 \\
			B-ORG & 2     & 0     & 0     & 2     & 0     & 0     & 0     & 1     & \textbf{0} \\
			\bottomrule
		\end{tabular}
	\end{center}
	
	\caption{Confusion matrix for NER on CoNLL'03 for the BILSTM with DP ($\varepsilon = 0.00837$). It can be seen that with this much privacy the model only randomly chooses tags (highlighted).}
	\label{tab:smallEps}
\end{table*}

\end{document}